\documentclass[runningheads]{llncs}

\usepackage{eccv}

\usepackage{eccvabbrv}

\usepackage{graphicx}
\usepackage{booktabs}

\usepackage[accsupp]{axessibility}  % Improves PDF readability for those with disabilities.

\usepackage{xcolor,colortbl}

\usepackage[pagebackref,breaklinks,colorlinks,citecolor=eccvblue]{hyperref}
\usepackage{orcidlink}

\usepackage{amsmath}
\usepackage{amssymb}
\usepackage{mathtools}
\usepackage[capitalize,noabbrev]{cleveref}

\usepackage{tabularx}

\definecolor{LightCyan}{rgb}{0.88,1,1}
\newcolumntype{a}{>{\columncolor{LightCyan}}c}

\usepackage{wrapfig}
\usepackage[normalem]{ulem}
\usepackage{ifthen}
\usepackage{amsmath}
\usepackage{adjustbox}
\usepackage{enumitem}

\newcommand{\tali}[2][]{%
    \ifthenelse{\equal{#1}{} }
        {\textcolor{magenta}{(TD) #2}}
        {\textcolor{magenta}{(TD) \sout{#1}\xspace{}#2}}
}

\newcommand{\narek}[2][]{%
    \ifthenelse{ \equal{#1}{} }
        {\textcolor{blue}{(NT) #2}}
        {\textcolor{blue}{(NT) \sout{#1}\xspace{}#2}}
}
\newcommand{\assaf}[2][]{%
    \ifthenelse{ \equal{#1}{} }
        {\textcolor{cyan}{(AS) #2}}
        {\textcolor{cyan}{(AS) \sout{#1}\xspace{}#2}}
}
\newcommand{\shai}[2][]{%
    \ifthenelse{ \equal{#1}{} }
        {\textcolor{orange}{(SB) #2}}
        {\textcolor{orange}{(SB) \sout{#1} #2\xspace{}}}
}
\usepackage{xcolor}
\usepackage{bbm}
\usepackage{xspace}
\usepackage{caption}
\newcommand{\mathtracker}{\Pi}

\newcommand{\mathrefinertensor}{\mathbf{\Phi}_{\Delta}}

\newcommand{\mathdino}{\mathbf{\Phi}_{\texttt{DINO}}}

\newcommand{\mathfeat}{\mathbf{\Phi}}
\newcommand{\mathfeatp}[1]{\pmb{\varphi}_{#1}}
\newcommand{\mathdinofeat}{\pmb{\varphi}_{\texttt{DINO}}}
\newcommand{\mathfeatt}[1]{\mathfeat^{#1}}

\newcommand{\mathframet}[1]{\mathbf{I}^{#1}}
\newcommand{\mathframe}{\mathbf{I}}

\newcommand{\wdinobb}{w_{\texttt{dino-bb}}^{ij}}
\newcommand{\wrefbb}{w_{\texttt{rfn-bb}}^{ij}}

\newcommand{\mathnn}{NN}

\newcommand{\mathcossim}{\text{cos-sim}}

\newcommand{\cmap}{\mathbf{S}}
\newcommand{\hmap}{\mathbf{H}}

\newcommand{\bsx}[1]{\mathbf{x}_{#1}}
\newcommand{\bp}[1]{\mathbf{#1}}

\newcommand{\plpl}{\raisebox{2pt}{\tiny ++}}

\newcommand{\mathflow}{\mathbf{f}}

\newcommand{\med}{\text{med}}
\newcommand{\flowij}{\bsx{}^{i \rightarrow j}}

\newcommand{\lossflow}{\mathcal{L}_{\texttt{flow}}}

\newcommand{\lossdinobb}{\mathcal{L}_{\texttt{dino-bb}}}
\newcommand{\lambdadinobb}{\lambda_1}

\newcommand{\losstunedbb}{\mathcal{L}_{\texttt{rfn-bb}}}
\newcommand{\lambdatunedbb}{\lambda_2}

\newcommand{\losscyc}{\mathcal{L}_{\texttt{rfn-cc}}}
\newcommand{\lambdacyc}{\lambda_3}

\newcommand{\lossref}{\mathcal{L}_{\texttt{rfn}}}

\newcommand{\lossnorm}{\mathcal{L}_{\texttt{norm}}}

\newcommand{\lossangle}{\mathcal{L}_{\texttt{angle}}}

\newcommand{\lossprior}{\mathcal{L}_{\texttt{prior}}}
\newcommand{\lambdaprior}{\lambda_4}

\newcommand{\lossclr}{l}

\newcommand{\lossnormdef}{\left| 1 - \frac{|| \mathfeat(\mathframe)[\bp{p}] ||_2}{|| \mathdino(\mathframe)[\bp{p}] ||_2} \right|}

\newcommand{\lossangledef}{\left| 1 - \mathcossim \left( \mathfeat(\mathframe)[\bp{p}], \mathdino(\mathframe)[\bp{p}] \right) \right|}

\newcommand{\dinotracker}{DINO-Tracker\xspace}

\newcommand{\dino}{DINO\xspace}
\newcommand{\afterfigure}{\vspace{-5mm}}  % -0.25cm}}
\newcommand{\featureres}{Delta-DINO\xspace}

\begin{document}

\title{\dinotracker: Taming DINO for Self-Supervised Point Tracking in a Single Video} 

\titlerunning{Taming DINO for Self-Supervised Point Tracking in a Single Video}

\author{ Narek Tumanyan$^*$,
Assaf Singer$^*$,
Shai Bagon,
Tali Dekel }

\authorrunning{N.~Tumanyan, A.~Singer et al.}

\institute{Weizmann Institute of Science \\
{\small *Indicates equal contribution.} \\
{\small Project webpage: \href{https://dino-tracker.github.io}{dino-tracker.github.io}}}

\maketitle

% ---------------------------------------------------------------
% Teaser
\begin{centering}
\label{fig:teaser}
\includegraphics[width=1\textwidth]{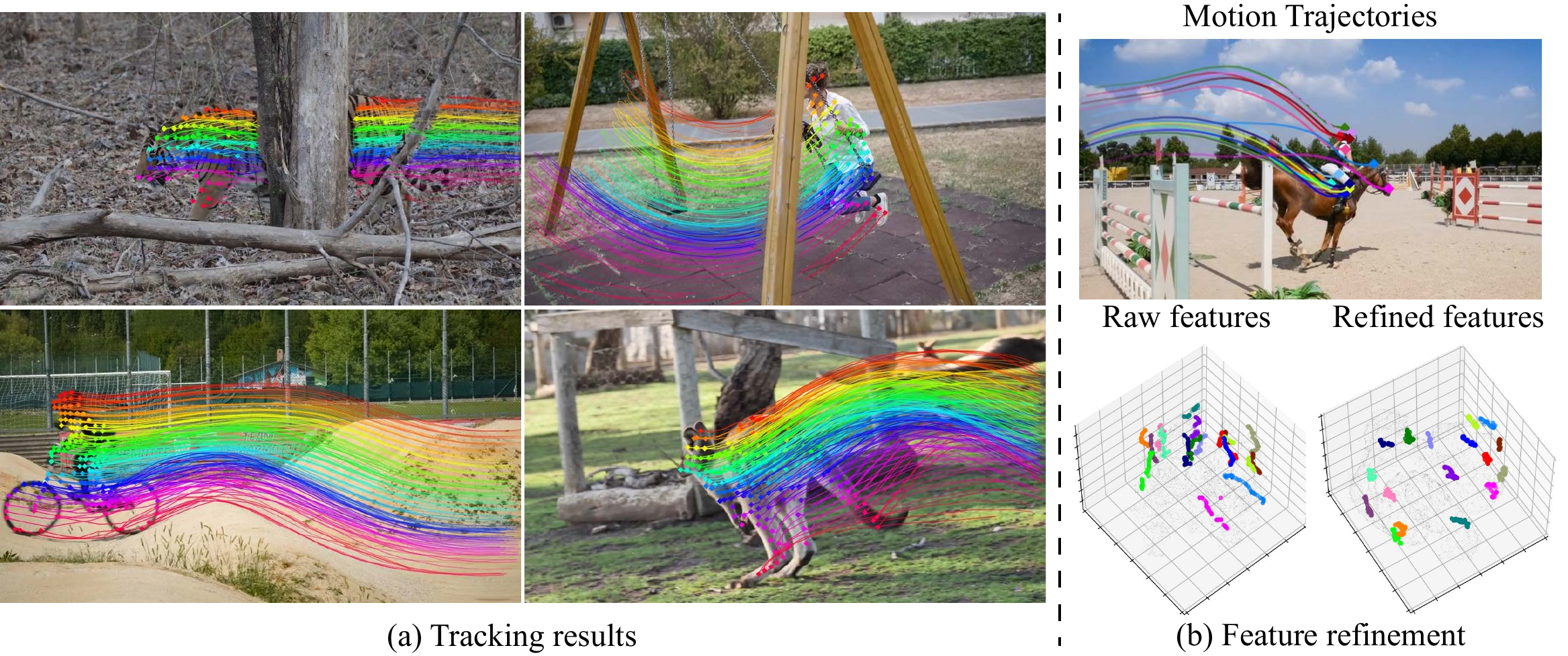} 
\captionof{figure}{
\emph{DINO-Tracker} provides long-range dense trajectories, past repeating occlusions and during challenging object deformations (a); For visualization purposes, the trajectories are shown for sampled points, yet our method tracks any point.  Our test-time training framework leverages a pre-trained DINO-ViT model, and optimizes its internal features for tracking in a single video. (b) \emph{Visualization of trajectory features using t-SNE:} We reduce the dimensionality of foreground features extracted from all frames to 3D using t-SNE, for both raw DINO features and our optimized ones; Features sampled along ground-truth trajectories are marked in color, where each color indicates a different trajectory. Our refined features exhibit tight ``trajectory-clusters'',  allowing our method to associate matching points across distant frames  and occlusion. 
}
\end{centering}

% ---------------------------------------------------------------
\begin{abstract}
We present \dinotracker~-- a new framework for long-term dense tracking in video. The pillar of our approach is combining test-time training on a single video, with the powerful localized semantic features learned by a pre-trained DINO-ViT model. Specifically, our framework simultaneously adopts DINO's features to fit to the motion observations of the test video, while training a tracker that directly leverages the refined features. The entire framework is trained end-to-end using a combination of self-supervised losses, and regularization that allows us to retain and benefit from DINO's semantic prior. Extensive evaluation demonstrates that our method achieves state-of-the-art results on known benchmarks. DINO-tracker significantly outperforms self-supervised methods and is competitive with state-of-the-art supervised trackers, while outperforming them in challenging cases of tracking under long-term occlusions.
\end{abstract}

\section{Introduction}
\label{sec:intro}
Establishing dense point correspondences in video has seen tremendous progress in recent years. In the case of short-term dense motion estimation, i.e., optical flow estimation,  the research community has been primarily focused on \emph{supervised learning} -- designing powerful feedforward models that are trained on various synthetic datasets, using ground truth accurate supervision~\cite{zhai2021optical}. Recently, this trend has been expanded to \emph{long-range} point tracking in video.   With the rise of new architectures (e.g., Transformers \cite{dosovitskiy2021an}) and new synthetic datasets that provide long-term trajectories supervision \cite{doersch2022tapvid, zheng2023point}, various supervised trackers have been developed, demonstrating impressive results~\cite{doersch2022tapvid,doersch2023tapir,karaev2023cotracker}.  Nevertheless, tracking \emph{every} point in a video across its \emph{entire temporal duration} poses fundamental challenges to this prevalent supervised approach. First, synthetic datasets for point tracking, which often consist of moving objects in unrealistic configurations, are limited in their diversity and scale, relative to the vast distribution of motion and objects in natural videos. In addition, existing models are still restricted in their ability to aggregate information across the entire spatiotemporal extent of a video -- a pivotal component in tracking especially under long-term occlusions (e.g., correctly matching a point before it is occluded and after it is revealed).    

Aiming to tackle the above challenges, Omnimotion~\cite{wang2023omnimotion} recently proposed to take the opposite direction through a test-time optimization framework that lifts tracking into 3D, and leverages pre-computed optical flow and video reconstruction as supervision. By optimizing a tracker on a given test video, this approach essentially solves for the motion of all video pixels at once. Nevertheless, a main drawback of Omnimotion is that it heavily relies on pre-computed optical flow and the information available in a \emph{single} video -- it does not benefit from \emph{external} knowledge and priors about the visual world. 

In this paper, we propose to close the gap between test-time training and learning from extensive data by combining the best of both worlds: a test-time optimization framework that is tailored to a specific video, coupled with the powerful feature representation learned by an external image model trained on broad unlabeled images. Specifically, inspired from the tremendous recent progress in self-supervised learning, our framework leverages a pre-trained DINOv2 model~\cite{oquab2023dinov2} -- a Vision Transformer distilled using a large collection of natural images. DINO's features have been shown to capture fine-grained semantic information and has been used for various visual tasks such as segmentation and semantic correspondences (e.g., \cite{amir2021deep, Shtedritski_2023_ICCV, MelasKyriazi2022DeepSM}). Our work is the first to consider these features for dense tracking. We show that using raw DINO feature matching can serve as a strong baseline for tracking, yet the features are not discriminative enough to support sub-pixel accurate tracking on their own, as can be seen in the t-SNE visualization of Fig.~\ref{fig:teaser}(b). Our framework simultaneously refines DINO's features to fit to the motion observations of the test video, while training a  tracker that directly leverages the refined features. 
To this end, we formulate a new objective function that goes beyond optical flow supervision by  fostering robust semantic feature-level correspondences derived from DINO within our refined feature space.

We extensively evaluate our framework across established benchmarks and 
demonstrate its superiority in scenarios requiring semantic understanding, dealing with appearance ambiguity, and handling long occlusions. Our tracker achieves state-of-the-art performance compared to previous self-supervised methods, and reveals a significant boost in tracking through long occlusion,  compared to  state-of-the-art supervised trackers.

~~To summarize, our contributions are as follows:
\begin{itemize}[leftmargin=*,topsep=0pt]
    \item We are the first method to harness pre-trained DINO features for point-tracking.
    \item We present the first method that combines test-time training with external priors for tracking.  
    \item We achieve a notable performance boost w.r.t. prior methods  in tracking through long-term occlusions.
\end{itemize}

\section{Related Work}\label{sec:related-work}
\paragraph{Optical flow.} 
Classical optical flow optimization methods are based on color constancy and motion smoothness (e.g.,~\cite{lucaskanade, Black1993AFF, HORN1981185, bruhn2005lucas_horn}). Later, these hand-crafted priors have been replaced by data-driven approaches (e.g.,~\cite{Dosovitskiy2015FlowNetLO, Ilg2016FlowNet2E, Sun2017PWCNetCF, raft, Huang2022FlowFormerAT, Xu2021GMFlowLO,xu2017DCFlow}), where modern deep learning-based optical flow methods typically  take a supervised learning approach by leveraging synthetic training data containing ground truth optical flow labels. 
While optical-flow estimation has seen great progress, establishing accurate dense correspondences between nearby frames, extending it to long-term tracking (e.g., by chaining pairwise correspondences) is hampered by occlusions and prone to error accumulation. In our method, we use RAFT~\cite{raft} to derive short-term motion supervision for our model.

\paragraph{Learning correspondences from videos} 
While optical flow focuses on dense matches between consecutive frames, other methods were developed for matching corresponding points across distant frames. 
Classical methods used hand-crafted features (e.g, \cite{lowe2004sift,Liu2011SIFTFD}), while
more recently, these correspondences were learned in a weakly or self-supervised manner~\cite{bian2022_contrastive_rw,caron2021emerging,li2019joint_ssl_temp_corr,rocco2018neighbourhood,Vondrick_2018_colorization,wang2020corr_pose,xu2021rethinking}. Some of these methods exploit video data to learn correspondences, using  various cues such as cycle-consistency in time~\cite{CVPR2019_CycleTime, jabri2020walk,zhou2016learning}. Nevertheless, at test time, these models operates on a pair of frames, and do not consider wider temporal context, which makes them unsuitable for dense point tracking.  

\paragraph{Feedforward models for dense tracking.} 
Recently, there has been notable progress in developing feedforward neural network-based models for dense tracking  (e.g., \cite{karaev2023cotracker, doersch2023tapir, doersch2022tapvid, zheng2023point, harley2022particle, neoral2024mft}). This advancement has been facilitated by the rise of new architectures and synthetic datasets that provide ground truth trajectory supervision \cite{doersch2022tapvid, zheng2023point}.   TAP-Net \cite{doersch2022tapvid} estimates the position of a query point by computing a cost volume for each target frame independently, followed by regressing the cost volume to a 2D coordinate and a visibility score. PIPs~\cite{harley2022particle} revisits classical particle-based representation \cite{Sand2006ParticleVL} by designing an MLP-based tracker that predicts tracklets in 8-frame window. To predict long-range tracks, PIPs is applied in a sliding-window fashion -- an approach that is prone to drifting errors and cannot handle long-term occlusions.  Aiming to extend the temporal field of view, PIPs\plpl \cite{zheng2023point} replaces the MLP-Mixer with a fully-convolutional 1D architecture. However, trajectories of different points are still predicted independently. Co-Tracker \cite{karaev2023cotracker} aims to tackle this issue through a new Transformer-based architecture that jointly tracks multiple query points, and demonstrated impressive results on several benchmarks such as TAP-Vid-DAVIS. However, their temporal field of view is still limited due to the expensive attention modules.  
TAPIR \cite{doersch2023tapir} combines  TAP-Net and PIPs design in a two-stage framework: first, tracks are initialized using per-frame cost volume estimation, which are then refined similarly to \cite{harley2022particle}.  Our work takes a different route in two fundamental ways: (i) all these methods are  \emph{trained from scratch} in a supervised manner. In contrast, we aim to leverage the rich and powerful internal representation learned by an external self-supervised image model, (ii) due to computational and memory requirements, these models are still limited in either their temporal or spatial field of view. We aggregate information across \emph{all} video pixels 
via the trained weights of the tracker which is optimized to a specific video.

Recently, \cite{sun2024refining} proposed a self-supervised scheme for improving pre-trained supervised motion estimation models, by self-distilling cycle-consistent predictions. However, their method relies solely on the pre-trained model and does not consider any external priors, which is the focus of our approach.

\paragraph{Optimization-based tracking.}
The task of long-term tracking dates back to classical works that optimize motion globally  over a video (e.g.,~\cite{Sand2006ParticleVL, Zhao2022ParticleSfMED, Rubinstein2012TowardsLL, Chang2013AVR}). However, these methods are restricted to sparse or semi-dense tracking, and struggle to track under occlusions. Recently, Omnimotion \cite{wang2023omnimotion} proposed a neural-based framework that performs tracking by learning a bijective mapping between each point in the video and a canonical quasi-3D space. Their model is optimized per-video in a self-supervised manner, using pre-computed optical flow and video reconstruction as supervision. Similarly, our method takes a test-time training approach, yet fundamentally differs from~\cite{wang2023omnimotion} in utilizing an external visual prior. As a result, \dinotracker outperforms Omnimotion in scenarios where reliable optical flow is lacking, such as tracking past long occlusions. Moreover, our optimization process is more time-efficient as we only refine \emph{pre-trained} features with a lightweight architecture.

\paragraph{DINO-ViT Features as local semantic descriptors.}
\dino~\cite{caron2021emerging} features were shown to effectively serve as dense and localized visual descriptors~\cite{amir2021deep} for many tasks such as finding semantic correspondences~\cite{amir2021deep, Shtedritski_2023_ICCV, Mariotti2023ImprovingSC, zhang2023telling, zhang2023tale}, performing segmentation and part-segmentation~\cite{amir2021deep, MelasKyriazi2022DeepSM, Aflalo2022DeepCutUS, hamilton2022unsupervised}, transferring appearance in a semantically aware manner\cite{tumanyan2022splicing, tumanyan2023disentangling}, and aligning a set of semantically related images -- establishing dense correspondences between them~\cite{ofriamar2023neural, gupta2023asic}. 
Recently, Time-tuning~\cite{salehi2023time} took \dino features to the temporal domain to improve the consistency of video segmentation.
Our work is the first to harness the semantic prior of \dino for the task of dense, sub-pixel, long-range tracking in video.

\section{Method}
Given an input video $\{ \mathframet{t} \}_{t=1}^{T}$, our goal is to train a tracker $\mathtracker$ that takes a query point $\bsx{\bp{q}}$ as input and outputs a set of position estimates $\{\hat{\bsx{}}^t\}_{t=1}^{T}$. As illustrated in Fig.~\ref{fig:pipeline}, our framework follows the prevailing approach of extracting features, for both the query $\bsx{\bp{q}}$ and a target frame $\mathframet{t}$, and estimating the final position $\hat{\bsx{}}^t$ based on the maximal location in the cost volume. The core of our method is harnessing a pre-trained DINOv2-ViT model \cite{oquab2023dinov2} in our feature extraction. DINO's pre-trained features provide our framework with an initial semantic and localized representation, yet, lacks temporal consistency and fine-grained localization required for accurate long-term tracking. We thus train \featureres~-- a feature extractor that predicts a residual to the pre-trained DINO features.

Our goal is to refine the features such that they can act as ``trajectory embeddings'', i.e., features sampled along a trajectory should converge to a unique representation, while preserving the original DINO prior. To this end, we formulate a new objective function that is used to train our tracker in a self-supervised manner, on a single input video. Our sources of supervision are: (i) pre-computed optical flow which provides us with pseudo ground truth short-term pixel-level correspondences, (ii) semantic feature-level correspondences extracted from raw DINO features, which are distilled into our refined feature space through a contrastive objective, and (iii) self-distillation losses  aiming to sharpen the correlation between reliable correspondences distilled from our refined feature space.
We next describe our tracking framework and supervision in detail.

\subsection{\dinotracker}
\label{sec:dino-tracker}
\begin{figure*}[t!]
  \centering
  \includegraphics[width=\linewidth]{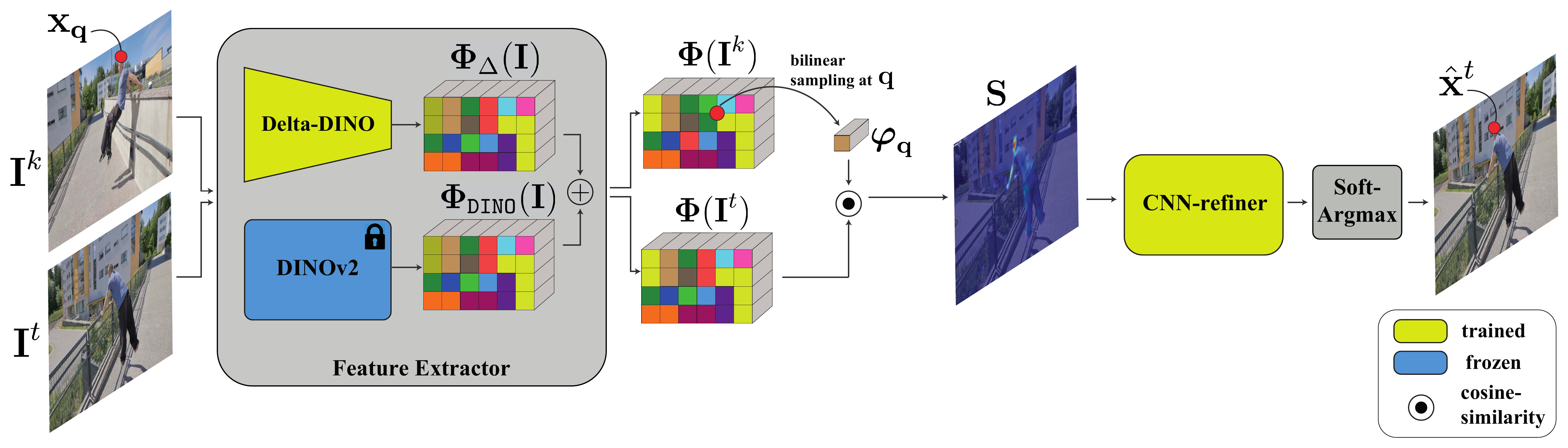}
  \caption{\emph{\dinotracker at inference:} Features are  extracted from a reference frame $\mathbf{I}^k$, and a target  frame $\mathbf{I}^t$. Our feature extractor consists of  a \emph{fixed} pre-trained DINOv2 model, and our CNN \featureres model, which predicts a residual to DINO's features. To track a query point $\bsx{q} \in \mathframet{k}$, we compute the cost volume between its sampled feature $\mathfeatp{q}$, and the target feature map  $\mathfeat{(\mathbf{I}^t)}$. The resulting  heatmap $\cmap$ is refined, and the final tracked position $\hat{\bsx{}}^t$ is estimated  based on points in the vicinity of the maximal location.}
  \label{fig:pipeline}
  % \afterfigure
\end{figure*}

The core component of our framework is the \featureres
model, which predicts the residuals to \emph{frozen} DINO features for frame $\mathframe$ (Fig.~\ref{fig:pipeline}). That is, our refined features $\mathfeat(\mathframe) \in \mathbb{R}^{H' \times W' \times C}$ are given by:
\begin{equation}
 \mathfeat(\mathframe) = \mathdino(\mathframe) + \mathrefinertensor(\mathframe)
\end{equation}
where $\mathdino(\mathframe)$ are the pre-trained DINO features, and $\mathrefinertensor(\mathframe)$ are the predicted residual features. We use a CNN-based model for \featureres, to benefit from its inductive bias, i.e., encoding similar RGB patches across frames into  similar feature representation. In addition, predicting a residual rather than directly fine-tuning DINO allows us to better retain its prior \cite{zhang2023adding}. To stabilize our fine-tuning process, we zero-initialize our refiner.

Given a query point $\bsx{\bp{q}}$ in  $\mathframe^k$, we bilinearly-sample its feature: $\mathfeatp{\bp{q}}= \mathfeat(\mathframe^k)[\bp{q}]$, where $\bp{q}$ is the rescaled coordinate of $\bsx{\bp{q}}$ in the feature map.  We then compute the cost volume between $\mathfeatp{\bp{q}}$ and a target feature map $\mathfeat^t=\mathfeat(\mathframe^t)$ as follows:

\begin{equation*}
    \cmap(\bp{p}) = \mathcossim(\mathfeatp{\bp{q}}, \mathfeatt{t}(\bp{p})) \quad \mbox{where} \quad \mathcossim(\mathbf{a}, \mathbf{b}) = \frac{\mathbf{a}^T \cdot \mathbf{b}}{|| \mathbf{a} ||_2 \cdot || \mathbf{b} ||_2}
\end{equation*}
Following \cite{doersch2022tapvid}, we input $\cmap$ to a small CNN-refiner network followed by a spatial softmax, resulting in the final heatmap $\hmap$. The final coordinate $\hat{\bsx{}}^t$ is computed by considering the points in the vicinity of the maximal location  $\bp{p}_{max}\in \hmap$ and computing their weighted sum:

\begin{equation}
    \hat{\bsx{}}^t = \frac{ \sum_{\bp{p} \in \Omega} \hmap(\bp{p}) \cdot \bsx{\bp{p}} } { \sum_{\bp{p} \in \Omega} \hmap(\bp{p}) }
    \label{eq:softargmax}
\end{equation}
where $\Omega = \{\bp{p}: || \bsx{\bp{p}} - \bsx{{\bp{p}}_{max}} ||_2 \leq R  \}$. Thus, the final output of our tracker is $\mathtracker(\bsx{\bp{q}}, t) = \hat{\bsx{}}^t$, and the track of $\bsx{\bp{q}}$ is $\mathcal{T}_q=\left\{\hat{\bsx{}}^t : \hat{\bsx{}}^t=\mathtracker(\bsx{\bp{q}}, t), t=1\ldots T\right\}$.

\subsection{Self-Supervision}\label{sec:supervision}
We train \dinotracker to match points along trajectories with supervising signals automatically extracted from the test video itself using RAFT optical flow and distilled feature correspondences.

\paragraph{Optical flow}
provides accurate, sub-pixel displacement information between consecutive frames. 
We extract short-term tracks by chaining these displacements over time.
A point $\bsx{}^i$ from frame $i$ is matched to $\bsx{}^j$ at frame $j$ if the optical flow tracklet between them is cycle-consistent. 
At preprocessing, we compute the set of all optical flow correspondences
$\Omega_{\texttt{flow}}=\left\{(\bsx{}^i, \bsx{}^j)\mbox{\ cycle-consistent}\right\}$, 
which provide high-quality supervision for short tracklets. However, they are not suitable for providing long-range supervision due to error accumulation (i.e. drifting) and occlusions. Further implementation details can be found in Appendix \ref{sec:preprocessing}.

\paragraph{Feature correspondences.}
are used to supplement our training data.
We extract feature correspondences from \dino and leverage them for additional supervision. 
Specifically, we extract reliable matches between pairs of feature maps $\mathdino(\mathframet{i}), \mathdino(\mathframet{j})$ by detecting ``best-buddy pairs'', i.e., mutual nearest neighbors~\cite{Dekel2015BestBuddiesSF}. Formally, a pair of points $\{\bp{p}^i, \bp{p}^j\}$ are best-buddies (bb) if:
\begin{equation}
    \mathnn(\mathdinofeat^{i}, \mathdino(\mathframet{j})) = \mathdinofeat^{j} \land \mathnn(\mathdinofeat^{j}, \mathdino(\mathframet{i})) = \mathdinofeat^{i}
    \label{eq:bbs}
\end{equation}
where $\mathnn(\mathfeatp{}, \mathfeat)$ is the nearest-neighbor of $\mathfeatp{}$ in feature map $\mathfeat$. 
At preprocessing, we compute the set of all \dino best-buddies 
$\Omega_{\texttt{dino-bb}}=\left\{\left(\bp{p}^i, \bp{p}^j\right)\mbox{\ DINO bb}\right\}$.

Additionally, during training, our refined features improve their representation and give rise to new reliable correspondences. We detect \emph{new} best buddies (Eq.~\ref{eq:bbs}) using the refined features, $\mathfeatp{}^i,\mathfeatp{}^j$. The set of refined best buddies, 
$\Omega_{\texttt{rfn-bb}}=\left\{\left(\bp{p}^i, \bp{p}^j\right)\mbox{\ refined bb}\right\}$, is constantly updated during training.

Importantly, these two sources of correspondences are complementary: while optical flow provides accurate \emph{sub-pixel} matches for near-by frames, features' best-buddies are extracted on a \emph{coarse} spatial grid but provide long-term matches. \dinotracker is optimized using both, enjoying the best of both worlds.

\subsection{Objective}\label{sec:objective}
Given an input video and the correspondences obtained in Sec.~\ref{sec:supervision}, we train our model using the following loss terms.

\paragraph{\bf Flow loss.}

To match our estimated tracks with the motion of the input video, we apply a flow loss $\lossflow$, which aligns the estimated positions with correspondences extracted from optical flow

\begin{equation*}
    \lossflow = \sum_{(\bsx{}^i, \bsx{}^j) \in \Omega_{\texttt{flow}}} L_H(\mathtracker(\bsx{}^i, j), \bsx{}^j) + L_H(\mathtracker(\bsx{}^j, i), \bsx{}^i)
\end{equation*}
where $\Omega_{\texttt{flow}}$ is the set of optical flow correspondences computed during preprocessing, and $L_H$ is Huber loss~\cite{Huber1964RobustEO}.

\paragraph{\bf DINO Best-Buddies Loss.}

Given a best-buddy pair $\{\bp{p}^i, \bp{p}^j\}\!\in\!\Omega_{\texttt{dino-bb}}$, we aim to increase the correlation between  their refined features  $\{\mathfeatp{}^{i},\mathfeatp{}^{j}\}$, while decreasing their correlation to other features using a contrastive loss~\cite{chen2020simple}: % \vspace*{-3mm}
\begin{equation*}
    \lossclr(\mathfeatp{}^{i}, \mathfeatp{}^{j}) = -\log{ \frac{ \exp( \mathcossim(\mathfeatp{}^{i}, \mathfeatp{}^{j}) / \tau ) }{ \sum_{\bp{p}} \exp( \mathcossim(\mathfeatp{}^{i}, \mathfeatt{j}(\bp{p}) ) / \tau  ) } }
\end{equation*}
where $\tau$ is a temperature parameter. Our DINO best-buddies loss is:

\begin{equation*}
    \lossdinobb = \frac{1}{|\Omega_{\texttt{dino-bb}}|} \sum_{(\mathfeatp{}^{i}, \mathfeatp{}^{j}) \in \Omega_{\texttt{dino-bb}}} \frac{1}{2}\wdinobb\left( \lossclr(\mathfeatp{}^{i}, \mathfeatp{}^{j}) + \lossclr(\mathfeatp{}^{j}, \mathfeatp{}^{i}) \right)
\end{equation*}

where $\wdinobb$ weights the loss for the corresponding pair based on a confidence metric of the detected best-buddy pair. The confidence is measured based on the unimodality of the similarity distribution between the pair of frames and on the correlation of the feature pair (see more details in Appendix~\ref{sec:apdx_training_details}).

\paragraph{\bf Refined Best-Buddies Loss.} 
We apply a similar contrastive loss for refined best-buddies distilled during training $\{\bp{p}^i, \bp{p}^j\} \in \Omega_{\texttt{rfn-bb}}$:

\begin{equation*}
    \losstunedbb = \frac{1}{|\Omega_{\texttt{rfn-bb}}|} \sum_{(\mathfeatp{}^{i}, \mathfeatp{}^{j}) \in \Omega_{\texttt{rfn-bb}}} \frac{1}{2}\wrefbb\left( \lossclr(\mathfeatp{}^{i}, \mathfeatp{}^{j}) + \lossclr(\mathfeatp{}^{j}, \mathfeatp{}^{i}) \right)
\end{equation*}
where $\wrefbb$ weights the loss for the corresponding pair based on the cosine-similarity of the features.

\paragraph{\bf Cycle-Consistency Loss.}
We also found it beneficial to encourage the preservation of cycle-consistent tracks produced by \dinotracker. A pair of points $\{ \bsx{}^i, \bsx{}^j \}$ is considered cycle-consistent if $\bsx{}^j = \mathtracker \left( \bsx{}^i, j \right)$ and $ || \mathtracker(\bsx{}^j, i) - \bsx{}^i ||_2 \leq \gamma $, where $\gamma$ is a small error threshold. Our cycle-consistency loss is given by:

\begin{equation}
    \label{eq:cyc}
    \losscyc = \sum_{(\bsx{}^i, \bsx{}^j) \in \Omega_{\texttt{rfn-cc}}} \frac{1}{2} w^{ij}_{\texttt{rfn-cc}} \left( L_H(\mathtracker(\bsx{}^i, j), \bsx{}^j) + L_H(\mathtracker(\bsx{}^j, i), \bsx{}^i) \right) 
\end{equation}

where $\Omega_{\texttt{rfn-cc}}$ are cycle-consistent coordinate pairs extracted during training, and $w^{ij}_{\texttt{rfn-cc}}$ weights each term according to the cycle-consistency error (see Appendix~\ref{sec:apdx_training_details} for details).

\paragraph{\bf Prior Preservation Loss.}
We apply regularization losses to preserve DINO's prior in our refined feature space: Specifically, we encourage each refined feature to: 1.~maintain a high cosine similarity, and 2.~have a close norm to its corresponding DINO feature. Given DINO features $\mathdino(\mathframe)$ and refined features $\mathfeat(\mathframe)$, our prior-preservation loss is defined as:

\begin{equation*}
    \lossprior = \frac{1}{H' \cdot W'} \cdot \sum_{\bp{p}}  \underbrace{\lossnormdef}_{\lossnorm} + \underbrace{\lossangledef}_{\lossangle}
\end{equation*}

Thus, our final objective is:
\begin{equation}
    \mathcal{L} = \lossflow + \lambdadinobb \lossdinobb 
    + \lambdatunedbb \losstunedbb + \lambdacyc \losscyc + \lambdaprior \lossprior
\label{eq:objective}
\end{equation}
where $\lambda_*$ sets the relative weights between the terms. We use a fixed set of $\lambda_*$ in all our experiments. See Appendix~\ref{sec:apdx_training_details} for further implementation details and Appendix~\ref{sec:apdx_complexity} for complexity details.

\subsection{Occlusion Prediction}

\begin{figure}[t!]
    \begin{minipage}[c]{.46\textwidth}
        \includegraphics[width=\linewidth]{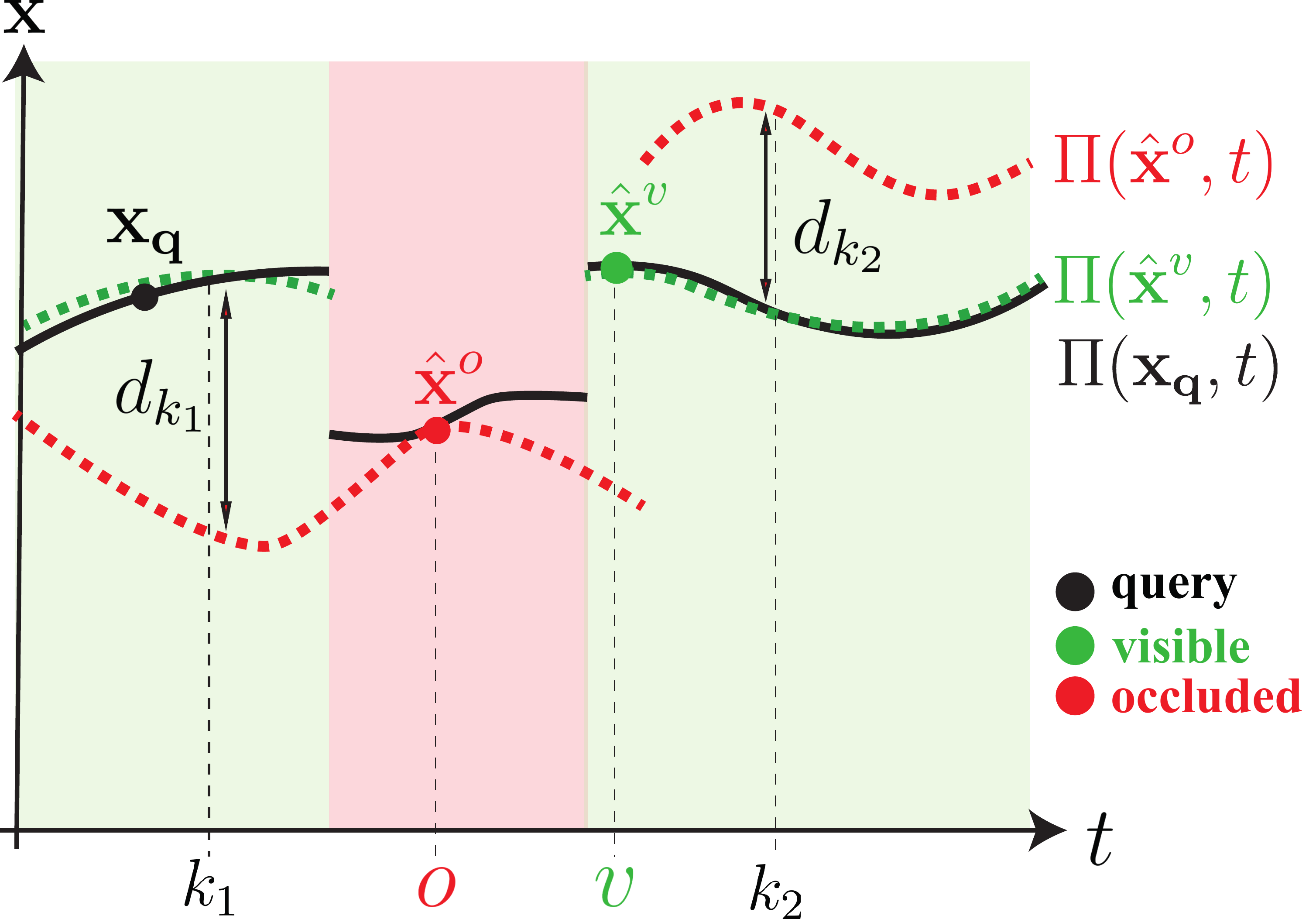} 
    \end{minipage}\hfill
    \begin{minipage}[c]{.54\textwidth}
        \caption{\emph{Visibility via trajectory agreement}. To determine the visibility of $\bsx{\bp{q}}$ at time $t\!=\!o$, we track $\hat{\bsx{}}^o$ across time and check the agreement between 
        $\mathtracker{(\hat{\bsx{}}^o, t)}$ and $\mathtracker{(\bsx{}, t)}$. This is done by measuring $d_{k_1},d_{k_2}$ -- displacements between the (black and red) tracks  for anchor time steps $k_1, k_2$.  Since these displacements are large, we classify $\bsx{\bp{q}}$ as occluded for $t\!=\!o$. For $t\!=\!v$, the track $\mathtracker{(\hat{\bsx{}}^v, t)}$ (green) agrees with $\mathtracker{(\bsx{}, t)}$, thus $\bsx{\bp{q}}$ is classified as visible for $t\!=\!v$.}
    \label{fig:occ-pred} 
    \end{minipage}
\end{figure}

Given an estimated trajectory $\mathcal{T}_\bp{q}$,
our goal is to determine if the query point $\bsx{\bp{q}}$ is indeed visible at each time $t$.
We do so based on trajectory agreement.
That is, if $\bsx{\bp{q}}$ is visible at time $t=v$, tracking from $\hat{\bsx{}}^v\in\mathcal{T}_{\bp{q}}$ will give rise to the same trajectory, i.e., $\mathtracker{(\bsx{\bp{q}},k)}\approx\mathtracker{(\hat{\bsx{}}^v, k)}$ for some frames $k$. 
This is illustrated by the agreement of the black $\mathcal{T}_{\bp{q}}$ and the green track in Fig.~\ref{fig:occ-pred}.
In contrast, if at time $t=o$ $\bsx{\bp{q}}$ is occluded, tracking from $\hat{\bsx{}}^o\in\mathcal{T}_{\bp{q}}$ will result with a different trajectory, i.e., $\left\|\mathtracker{(\bsx{\bp{q}},k)}-\mathtracker{(\hat{\bsx{}}^o, k)}\right\|=d_k$ will be large. 
This is illustrated by the red trajectory.
We measure this trajectory agreement on a few anchor frames $k=k_1,k_2,\ldots$ as illustrated in the figure.
To conclude, $\hat{\bsx{}}^t$ is deemed visible if $d_{k_1},d_{k_2},\ldots $ are small and the feature $\mathfeatp{}^t$ is similar to $\mathfeatp{\bp{q}}$.
More technical details on selecting anchor frames and various thresholds can be found in Appendix~\ref{sec:occ_pred_appendix}.

\section{Results}

\label{sec:results}

\paragraph{\bf Benchmarks.}
We evaluate our method on known benchmarks containing annotated trajectories on real videos:  (i)~\textbf{TAP-Vid-DAVIS}~\cite{doersch2022tapvid}, contains 30 object-centric videos of 34-104 frames, taken from \cite{davis}. (ii)~\textbf{TAP-Vid-Kinetics} contains 1189 videos of 250 frames each taken from \cite{kinetics}, depicting mostly human activity under both camera and objects' motion. We use the same set of 100 sampled videos used in \cite{wang2023omnimotion} for our evaluation. (iii)~\textbf{BADJA}~\cite{biggs2018badja}, contains 9 videos, at 480px resolution, depicting naturally moving animals with ground truth annotated keypoints.

\paragraph{\bf Metrics.}
The following metrics are measured for TAP-Vid benchmarks \cite{doersch2022tapvid}: 
\begin{itemize}[leftmargin=*,topsep=0pt]
    \item \emph{Position accuracy $\delta^x_{avg}$} measures the average position accuracy of visible points: $\delta^x_{avg} = \mathbb{E}_{x}(\delta^x)$, where each $\delta^x$ is the fraction of predicted points within the $x$ pixels neighborhood of the ground-truth position, where $x\!\in\!\left\{ 1, 2, 4, 8, 16 \right\}$.
    \item \emph{Occlusion Accuracy (OA)} measures the fraction of points with correct visibility prediction.
    \item \emph{Average Jaccard (AJ)} jointly measures position and occlusion accuracy.
\end{itemize}
The following metrics are used for evaluating BADJA:
\begin{itemize}[leftmargin=*,topsep=0pt]
    \item $\delta^{seg}$ measures the accuracy of the tracked keypoint within the distance of $0.2\sqrt{A}$ of the ground-truth annotation, where $A$ is the area of the foreground object in a frame.
    \item $\delta^{3px}$ measure the accuracy within a threshold of 3px.
\end{itemize}

\paragraph{\bf Baselines.}
We compare to state-of-the-art supervised feedforward trackers: PIPs\plpl \cite{zheng2023point}, TAP-Net~\cite{doersch2022tapvid}, TAPIR~\cite{doersch2023tapir} and Co-Tracker~\cite{karaev2023cotracker}, as well as the test-time optimization tracker Omnimotion~\cite{wang2023omnimotion}.

We consider two additional baselines: RAFT~\cite{raft}, in which tracking is performed by chaining optical flow displacements between consecutive frames, and  DINOv2~\cite{oquab2023dinov2}, using nearest neighbor matching between raw DINOv2 features. Since DINO features are computed at low resolution, the position in RGB space is obtained using a weighted sum around the nearest neighbor (Eq.~\ref{eq:softargmax}). See Appendix~\ref{sec:ablation_details} for implementation details. Since Omnimotion requires hours of training for each video, in Kinetics, we evaluate only on 256-resolution, where pre-trained weights are available.

\begin{table}[t!]
\newcommand{\first}[1]{\textbf{#1}}
\newcommand{\second}[1]{\underline{#1}}
\newcommand{\third}[1]{#1} 
  \caption{\emph{Quantitative comparison}. We compare our performance to all the baselines on TAP-Vid-DAVIS, TAP-Vid-Kinetics \cite{doersch2022tapvid} and BADJA \cite{biggs2018badja} using the metrics described in Sec.~\ref{sec:results}. Methods that do not predict occlusions lack OA and AJ. Our test-time self-supervised tracker performs on-par with SOTA supervised \cite{karaev2023cotracker,doersch2023tapir}, while substantially outperforming the SOTA test-time training method \cite{wang2023omnimotion}. Higher is better for all metrics.\label{tab:comparison}} 

  \begin{adjustbox}{max width=\textwidth}
  \begin{tabular}{c|c|c|c|c|c|c|c|c|c|c|c|c||c|c|}
    \toprule
     Method & \multicolumn{3}{c|}{ DAVIS-256} & \multicolumn{3}{c|}{DAVIS-480} & \multicolumn{3}{c|}{ Kinetics-256} & \multicolumn{3}{c||}{ Kinetics-480} & \multicolumn{2}{c|}{ BADJA} \\
     &  $\delta^{x}_{avg} $ &  OA &  AJ &  $\delta^{x}_{avg}$ & OA & AJ & $\delta^{x}_{avg}$ & OA & AJ & $\delta^{x}_{avg}$ & OA & AJ & $\delta^{\textit{seg}}$ & $\delta^{3px}$ \\
    \hline
    RAFT \cite{raft} &  56.7 & -- & -- & 66.7 & -- & -- & 50.4 & -- & -- &  60.5 & -- & -- &  45.0 & 5.8 \\
    DINOv2 \cite{oquab2023dinov2} & 61.4 & -- & -- & 64.7 & -- & -- & 60.3 & -- & -- & 61.0 & -- & --  & 62.8 & 8.4 \\
    \hline
    TAP-Net$^{\star}$ \cite{doersch2022tapvid} & 53.4 & 81.4 & 38.4 & 66.4 & 79.0 & 46.0 & 61.7  & 86.6 & 48.5 & 67.1 & 81.5 & 47.7 &  45.4 & 9.6 \\
    PIPs\plpl$^{\star}$ \cite{zheng2023point} & 71.5 & -- & -- & 73.6 & -- & -- & 68.2 & -- & -- & \third{70.8} & -- & -- & 59.0 & 9.8 \\
    TAPIR$^{\star}$ \cite{doersch2023tapir} & \third{74.7} & \first{89.4} & \second{62.8} & \third{77.3} & \first{89.5} & \first{65.7} & \third{69.5} &\second{89.1} & \third{57.3} &  69.8 &  \third{86.7} & \third{57.5} & \second{68.7} & \third{10.5} \\
    Co-Tracker$^{\star}$ \cite{karaev2023cotracker} & \first{79.2} &  \second{89.3} & \first{65.1} & \second{79.4} & \first{89.5} & \second{65.6} & \second{72.9} & \third{88.9} & \first{59.9} & \second{72.8} & \second{88.9} & \second{59.8} &  	\third{64.0} & \second{11.2} \\
    \hline
    Omnimotion$^{\dagger}$ \cite{wang2023omnimotion} & 67.5 & 85.3 & 51.7 & 74.1 & 84.5 & 58.4 & 69.2 & \first{89.2} & 55.0 &  -- & -- & -- & 45.2 & 6.9 \\    
    Ours$^{\dagger}$ & \second{78.2} & \third{87.5} & \third{62.3} & \first{80.4} & \second{88.1} & \third{64.6} & \first{73.3} & 88.5 & \second{59.7} & \first{74.3} & \first{89.2} & \first{60.9} &  \first{72.4} & \first{14.3} \\
    \bottomrule
\end{tabular}
\end{adjustbox}
{\scriptsize $^{\star}$ -- supervised. \quad $^{\dagger}$ -- test-time training.}
\end{table}

\begin{figure}
  \centering
  \includegraphics[width=1\textwidth]{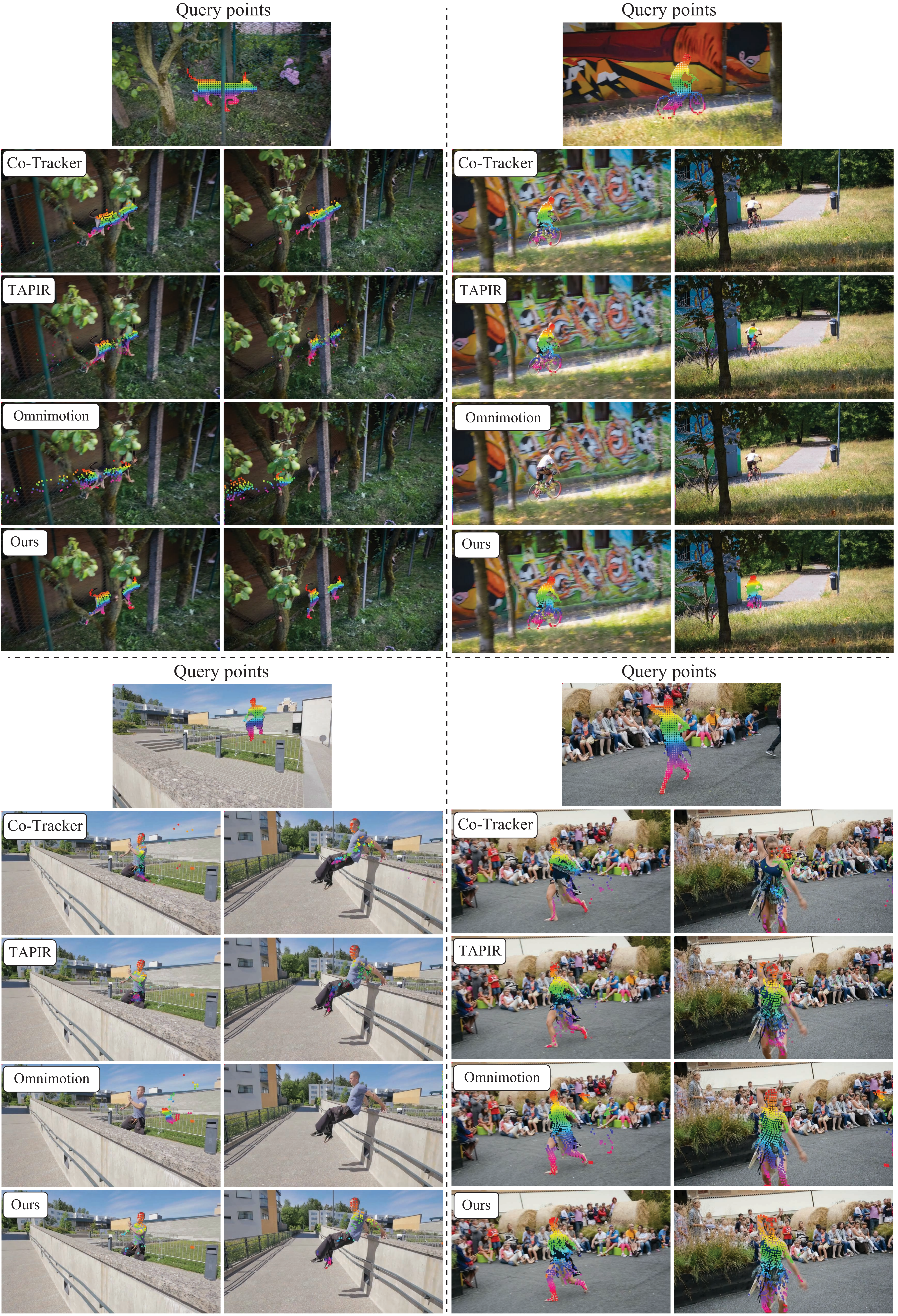}
  \caption{\emph{Qualitative results on TAP-Vid-DAVIS (480)}  Query points are color-coded on a reference frame (top). Our method exhibits better association of tracks across occlusions compared to SOTA trackers. Full videos and additional results are in the supplementary materials (SM) on our website.}
  \label{fig:comp-tapvid}
\end{figure}

\subsection{Comparisons} 
Table~\ref{tab:comparison} reports our performance on TAP-Vid benchmarks (for both 256px and 480px frame resolution) and BADJA (see Appendix \ref{sec:apdx_benchmarks} for details of evaluation).
As seen, raw DINOv2 is a surprisingly strong baseline: despite operating on low-resolution features, it outperforms RAFT, and even outperfroms TAP-Net, which is trained in a supervised manner for tracking, on DAVIS-256. Moreover, both RAFT and DINOv2 perform better on higher resolution.

Our method consistently outperforms all baselines on position accuracy ($\delta^x_{avg}$) on TAP-Vid, apart from Co-Tracker on DAVIS-256. Generally, all methods perform better on higher resolution. In our case, this is expected given the performance of raw DINOv2. 
Notably, compared to Omnimotion, which is the only test-time optimization competitor, our method exhibit a significant boost in performance across all benchmarks. This makes our method state-of-the-art among self-supervised baselines, and demonstrate the power of combining test-time training with external priors. In terms of our occlusion prediction (\emph{OA}), our performance is on-par with other methods, including supervised methods that use ground truth visibility labels. 

Figure.~\ref{fig:comp-tapvid} shows sample qualitative results on DAVIS-480. The objects in the top two videos are fast moving and are repeatedly occluded. As seen, all competitors struggle tracking through these occlusions, often tracking points to visually similar yet semantically unrelated regions (e.g. foreground points tracked to the background). 
Our results depict more semantically consistent trajectories. The bottom videos depict articulated objects and self-occlusion -- a particularly challenging scenario for all methods. Here too, our method tracks more persistently the foreground objects (e.g., head and upper-body of the man, woman's hands).

\begin{figure}[t]
  \centering
  \includegraphics[width=1\textwidth]{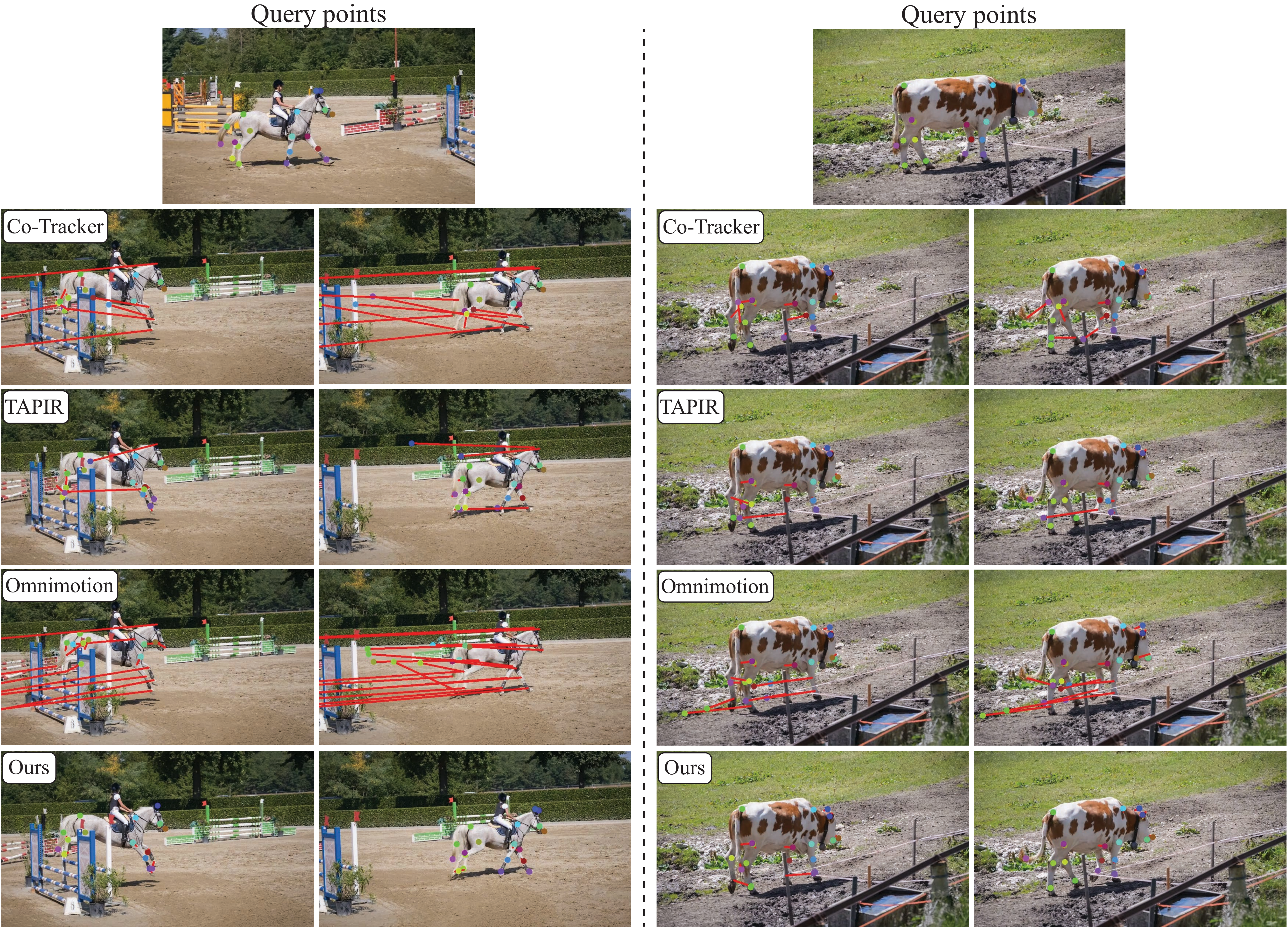}
  \caption{\emph{Sample results on BADJA w.r.t. ground truth.}
  Query points are color-coded on the frame at the top. Tracked points are marked on the target frames. Red lines indicate tracking \emph{errors} w.r.t. the ground truth positions.}
  \label{fig:comp-badja}
\end{figure}

Our results on BADJA, as seen in Table~\ref{tab:comparison}, are state-of-the-art in both $\delta^{seg}$ and $\delta^{3px}$ metrics. The positional accuracy w.r.t. ground truth is illustrated for sample examples in Fig.~\ref{fig:comp-badja}. 

\paragraph{\bf Tracking across occlusions.}
As discussed in Sec.~\ref{sec:supervision}, DINO's features provide complementary information to pixel-level optical flow, which allows our method to reason about correspondences across distant frames. This grants our method an advantage in tracking across long-term occlusions. To quantify this, we split TAP-Vid-DAVIS into three sets of videos with an increasing rate of occlusion. Specifically, for each trajectory, we compute the ratio of the number of occluded points to the length of the trajectory.

\begin{figure}[t]
  \centering
  \includegraphics[width=1\textwidth]{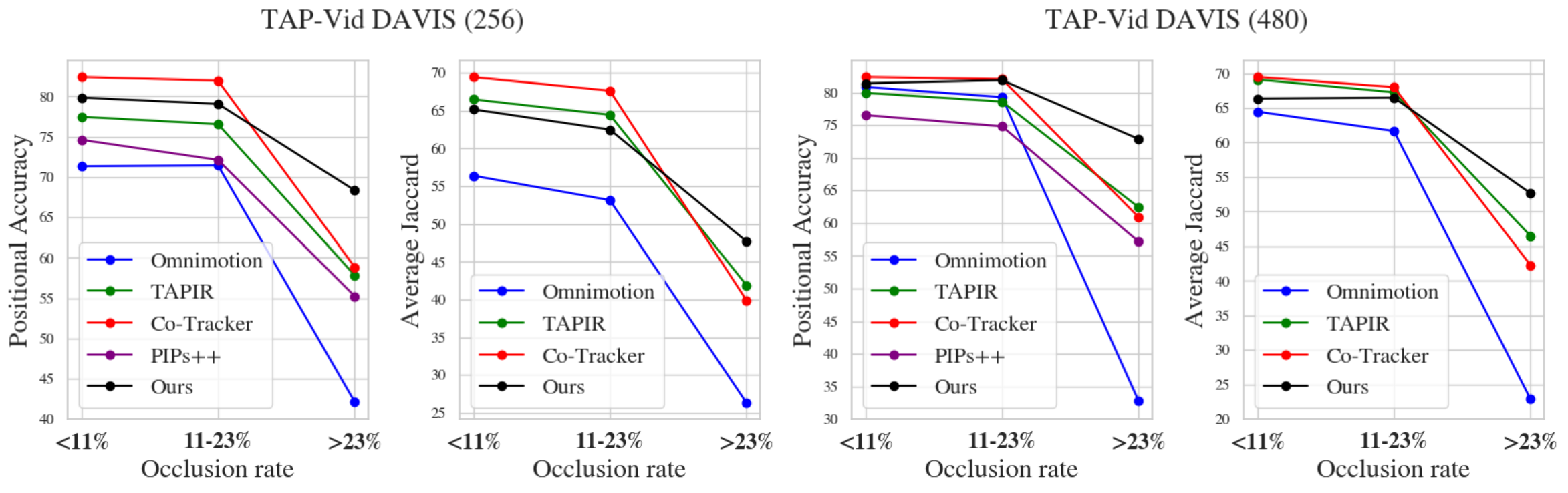}
  \caption{\emph{Tracking performance by occlusion rate.} 
  We group test videos from TAP-Vid DAVIS into three sets according to  occlusion rate (estimated using ground-truth visibility annotations). Positional accuracy and Average Jaccard are reported for each set separately.
  While the performance of all methods decreases as the occlusion rate increases, our \dinotracker exhibits a smaller gap and outperforms all methods with a large margin under a high occlusion rate. This demonstrates the benefit of harnessing the semantic information encoded in DINO's pre-trained features.  Omnimotion~\cite{wang2023omnimotion}, which solely relies on optical flow and video reconstruction, struggles in this case.
  }
  \label{fig:metrics_by_occ_davis}
\end{figure}

Figure~\ref{fig:metrics_by_occ_davis} reports the performance of our method and the baselines as a function of the occlusion rate. As seen, \dinotracker performs significantly better in case of a high occlusion rate due to the prior visual knowledge incorporated in the framework, enabling it to associate points across long-term occlusions.

\subsection{Ablations and Analysis} \label{sec:ablations}

\begin{figure*}[t!]
  \centering
  \includegraphics[width=1\textwidth]{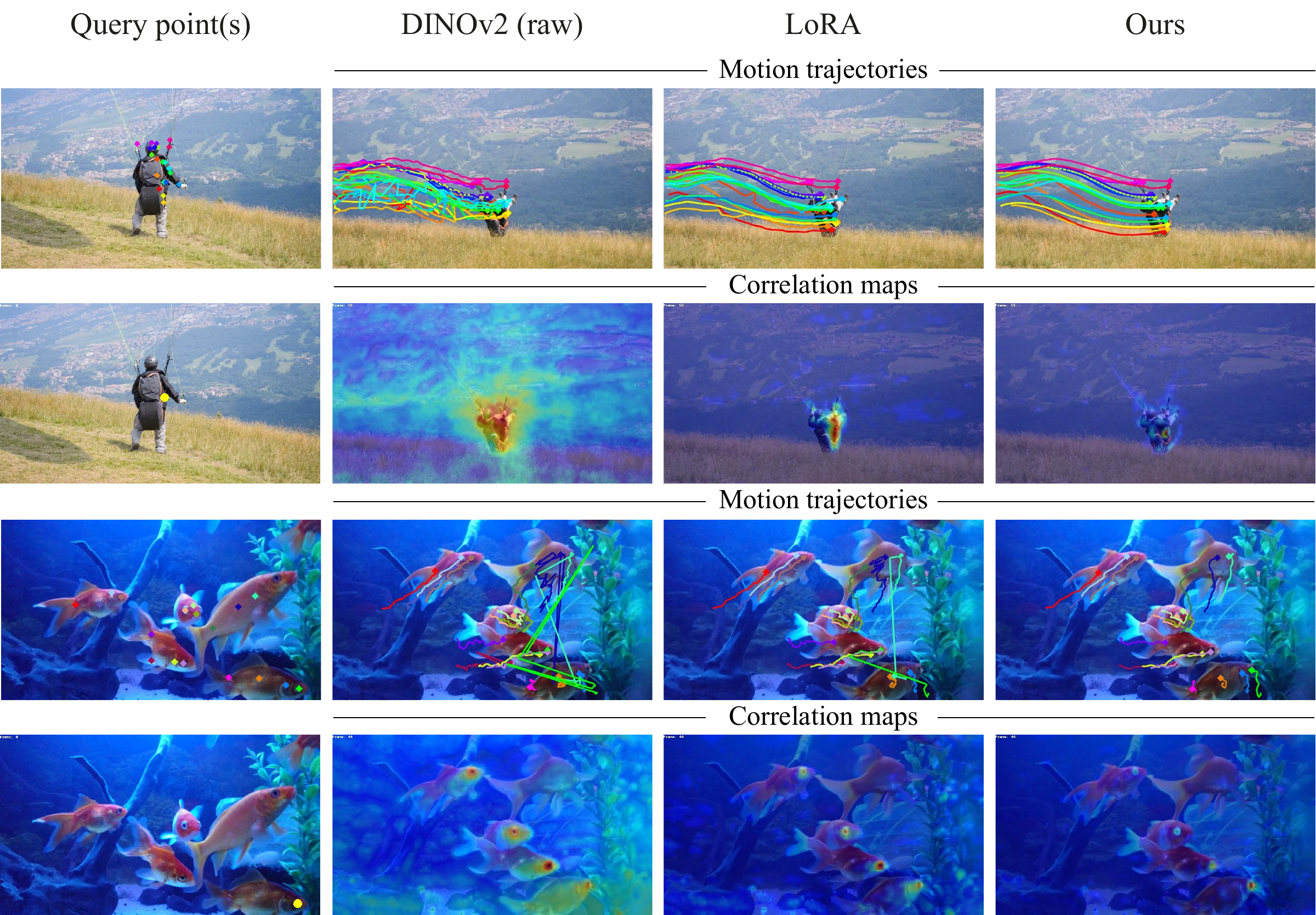}
  \caption{
  \emph{Comparing \dinotracker to (i) raw DINOv2 tracking, (ii) LoRA fine-tuning of DINOv2 for tracking}. 
  For each example, the top row shows color-coded query points and the corresponding tracks.
  The second row shows the correlation maps (cost volumes) between a single query point (marked in yellow) and all features of the target frame. Raw and LoRA features are not well localized and are ambiguous for semantically similar objects (e.g., eyes of the fish), yielding imprecise tracks. In contrast, our refined features are well localized and better resolve ambiguities.}
  \label{fig:ablation}
  \afterfigure
\end{figure*}

We quantitatively ablate our key design choices in Table~\ref{tab:ablation}. To quantify the contribution of DINO's prior,  we compare our full framework to a baseline in which $\mathdino(\mathframe) = \mathbf{0}$, i.e., we do not use \dino at all and train a CNN feature extractor from scratch, without $\lossprior, \lossdinobb$ losses in Eq.~\ref{eq:objective}. This baseline relies on appearance-based features only and performs dramatically worse in all metrics (\emph{w/o DINO} in Tab.~\ref{tab:ablation}).

We further consider a baseline in which our Delta-DINO CNN is replaced by fine-tuning DINOv2 weights using LoRA~\cite{hu2022lora}, using the same objective (Eq.~\ref{eq:objective}). As seen in Tab.~\ref{tab:ablation}, the performance significantly drops. We found that this approach produces  jittery trajectories, and that the heatmaps are less localized. This is seen in Fig.~\ref{fig:ablation} where we show the predicted tracks and correlation maps (cost volumes) for a couple representative examples. In contrast, our framework benefits from the inductive bias of CNN's as it learns to correlate similar RGB patches/neighborhoods, while also benefiting from the smoothness of CNN features. Another advantage of ours over LoRA is efficiency in memory and time.

In addition, Fig.~\ref{fig:ablation} includes the results of tracking based on \emph{raw} DINOv2 features. As seen, our optimization refines this initialization, leading to highly-localized heatmaps, even in  ambiguous regions (multiple fish eyes, paraglider body). This is also evident in Fig.~\ref{fig:teaser},  where we used t-SNE~\cite{van2008tsne} to visualize raw DINOv2 features and our refined features along \emph{ground-truth} tracks. DINOv2 features along trajectories are often ``spread out'' and are intertwined with features from other trajectories. In contrast, our refined features along a trajectory are distinctly clustered, making tracking more robust and accurate.

Finally, we quantify the contribution of each loss term in our objective (last rows of Tab.~\ref{tab:ablation}). Removing each term results in a drop in tracking performance and highlights their contribution. Interestingly, w/o $\lossflow$ reduces positional accuracy only by 2\%. This shows the effectiveness of combining DINO prior with our self-supervision and feature refinement for accurate tracking.

DINO features are the cornerstone of our framework. But which DINO features should we use? Tab.~\ref{tab:dino-feature} shows track position accuracy for different choices of DINOv2 ViT-L/14 facets. Using tokens extracted from the 16$^{th}$ layer performs the best, and we use these DINO features in all experiments.

\begin{table}[t]
%\begin{wraptable}{L}{0.5\textwidth}
\begin{minipage}{0.5\textwidth}
  \caption{\emph{Ablation study}. Removing one key component of our method at a time and reporting  performance on TAP-Vid-DAVIS videos. $\lossref$ is the combination of the losses $\losstunedbb$ and $\losscyc$.}
  \label{tab:ablation}
  \centering

  \begin{adjustbox}{max width=\textwidth}
  \begin{tabular}{c|c|c|c|}
    \toprule
     & \multicolumn{3}{c|}{ DAVIS-480} \\
     & $\delta^{x}_{avg}$ &  OA &  AJ \\
    \hline
    w/o \dino & 71.4 & 79.7 & 51.0 \\
    LoRA tune& 73.2 & 84.8 & 58.0 \\
    
    \hline

    w/o $\lossprior$ & 79.2 & 84.8 & 61.0 \\
    
    w/o $\lossref$ & 79.6 & 85.4 & 63.2 \\
    w/o $\lossdinobb$ & 78.2 & 87.0 & 61.9 \\
    w/o $\lossflow$ & 78.3 & 87.2 & 62.0 \\
    
    Ours & \textbf{80.4} & \textbf{88.1} & \textbf{64.6} \\
    \bottomrule
  \end{tabular}
  \end{adjustbox}
\end{minipage}\quad
\begin{minipage}{0.45\textwidth}
  \caption{\emph{DINO's feature layer ablation}. We evaluate  tracking performance using  DINOv2 ViT-L/14 features extracted from different layers and facets. We report track position accuracy ($\delta^x_{avg}$) on TAP-Vid-DAVIS 480. Based on these results we  use tokens extracted from the 16$^{th}$ layer.}
  \label{tab:dino-feature}
  \centering
  \begin{tabular}{c|c|c|c|c|}
    \toprule
     layer & tokens &  queries &  keys & values \\
     
    \hline
    12$^{th}$ & 61.1 & 51.1 & 50.0 & 62.4 \\
    16$^{th}$ & \textbf{64.7} & 48.8 & 46.7 & 63.9 \\
    20$^{th}$ & 63.8 & 56.9 & 56.0 & 64.0 \\
    23$^{rd}$ & 59.9 & 58.5 & 58.0 & 60.0 \\
    \bottomrule
  \end{tabular}

\end{minipage}

\end{table}

\section{Discussion and Conclusions}

We presented a new method for dense pixel-level tracking in video which combines test-time training on a single video with the power of external priors of a pre-trained DINO model. We introduced a new optimization-based framework that harnesses DINO's internal representation, while adapting it to the task of point tracking in a self-supervised manner. We demonstrated that our CNN-based design provides implicit smoothness prior effective for tracking. We demonstrated that our CNN-based design effectively preserves DINO's prior and provides implicit smoothness prior. 

Regarding limitations, while our method excels in associating points \emph{across} long-term occlusions, we do not model point trajectories \emph{behind} occluders. Previous methods achieve this using synthetic data for supervision, or lifting tracking into 3D.
However, a simple interpolation technique such as cubic spline can give plausible tracks during occlusion (see our SM website for examples).
Furthermore, our performance depends on the information encoded in DINO's pre-trained features. We observed that in challenging videos for which there is almost no optical flow supervision and there are multiple semantically-similar objects, trajectories may jump from one object to another. This is because raw DINO is mostly dominated by semantic information. 

We demonstrated the strengths of our \dinotracker through extensive evaluation and showed its superiority in associating points across long-term occlusions.  We hope that our work will trigger more research in leveraging self-supervised representation learning for dense tracking in video.

\subsection*{Acknowledgements}
We would like to thank Rafail Fridman for his insightful remarks and assistance. We would also like to thank the authors of Omnimotion for providing the trained weights for TAP-Vid-DAVIS and TAP-Vid-Kinetics videos.
The project was supported by an ERC starting grant OmniVideo (10111768), by Shimon and Golde Picker, and by the Carolito Stiftung.

Dr. Bagon is a Robin Chemers Neustein AI Fellow. He received funding from the Israeli Council for Higher Education (CHE) via the Weizmann Data Science Research Center and MBZUAI-WIS Joint Program for AI Research.

\clearpage
\newpage

\bibliographystyle{splncs04}
\bibliography{main}

%%%%%%%%%%%%%%%%%%%%%%%%%%%%%%%%%%%%%%%%%%%%%%%%%%%%%%%%%%%%%%%%%%%%%%%%%%%%%%%%%%%%%
%%%%%%%%%%%%%%%%%%%%%%%%%%%%%%%%%%%%%%%%%%%%%%%%%%%%%%%%%%%%%%%%%%%%%%%%%%%%%%%%%%%%%
%%%%%%%%%%%%%%%%%%%%%%%%%%%%%%%%%% Appendix %%%%%%%%%%%%%%%%%%%%%%%%%%%%%%%%%%%%%%%%%
%%%%%%%%%%%%%%%%%%%%%%%%%%%%%%%%%%%%%%%%%%%%%%%%%%%%%%%%%%%%%%%%%%%%%%%%%%%%%%%%%%%%%
%%%%%%%%%%%%%%%%%%%%%%%%%%%%%%%%%%%%%%%%%%%%%%%%%%%%%%%%%%%%%%%%%%%%%%%%%%%%%%%%%%%%%

\appendix
\newpage

\section{Implementation Details}

\subsection{Preprocessing}\label{sec:preprocessing}

\paragraph{Optical flow.} As discussed in Sec.~\ref{sec:supervision}, our method chains RAFT optical flow~\cite{raft} between consecutive frames, forming short-term accurate tracks for supervision. Specifically, for a given point $\bsx{}^{i}$ in frame $\mathframet{i}$, we generate a tracklet  $\{ \bsx{}^{j} = \bsx{}^{j-1} + \mathflow_{j-1 \rightarrow j}(\bsx{}^{j-1}); j \in \{ i+1, ..., t \} \}$, where $\mathflow_{j-1 \rightarrow j}$ is the optical flow between frames $\mathframet{j-1}$ and $\mathframet{j}$. We terminate the track at a frame $t$ if $|| \bsx{}^{t} - (\bsx{}^{t+1} + \mathflow_{t+1 \rightarrow t}(\bsx{}^{t+1})) || \geq \gamma_{\texttt{of}}$, where $\gamma_{\texttt{of}} = 1.5$px is a cycle-consistency threshold. To avoid drift error, we apply cycle-consistency checks on optical flow between distant frames. That is, we filter-out correspondences $\bsx{}^{j}$ that are inconsistent with $\mathflow_{i \rightarrow j}$, i.e. if
$|| \bsx{}^{j} - \flowij ||_2 \geq \gamma_{\texttt{of-lng}}$ and
$|| \bsx{}^{i} - ( \flowij + \mathflow_{j \rightarrow i}( \flowij ) ) ||_2 \leq \gamma_{\texttt{of}}$,
where $\flowij = \bsx{}^{i} + \mathflow_{i \rightarrow j}(\bsx{}^{i})$, $\gamma_{\texttt{of-lng}} = 2$px, and the second condition ensures that $\flowij$ is reliable. For each frame $\mathframet{i}$, we initialize new tracklets for points that do not have correspondences. The set of all correspondences processed from the optical flow is denoted as $\Omega_{\texttt{flow}} = \{ (\bsx{}^i, \bsx{}^j) \}$. In all our losses, continuous coordinates are being normalized to $[-1, 1]$.

\begin{figure}
  \centering
  \includegraphics[width=1\textwidth]{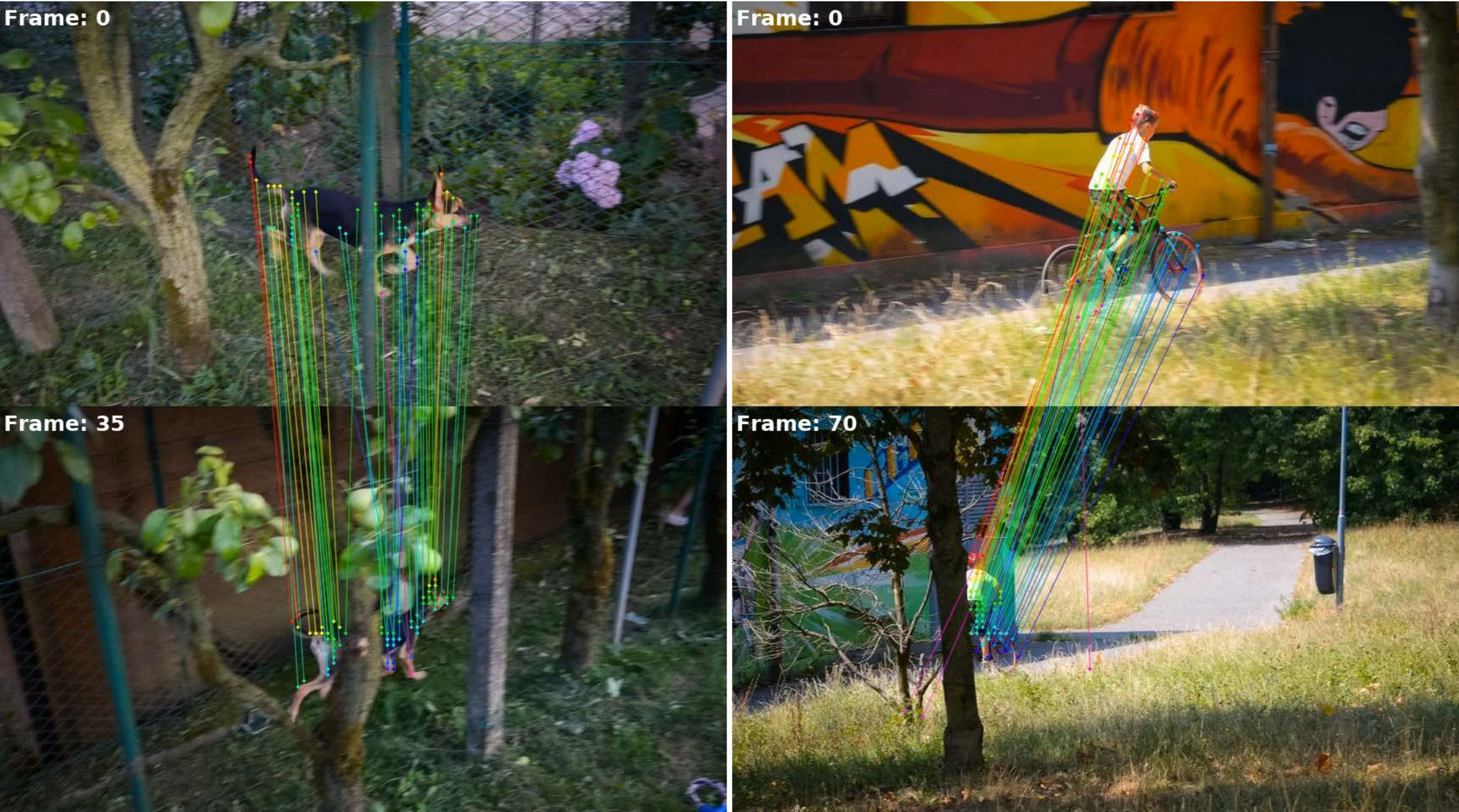}
  \caption{\emph{DINO best-buddies.} 
  We visualize best-buddy pairs between distant frames. DINO best-buddies provide localized semantic correspondences, allowing the model to recover the object past repeating occlusions.
  }
  \label{fig:dino_bb}
\end{figure}

\paragraph{DINO feature correspondences.} Since the \emph{coarse} feature correspondence supervision complements the \emph{sub-pixel} optical flow supervision, we discard feature correspondences for which optical flow supervision is available: $\Omega_{\texttt{dino-bb}} = \{ (\bp{p}^i, \bp{p}^j) \mbox{\ DINO best-buddy}: (\bsx{}^i, \bsx{}^j) \notin \Omega_{\texttt{flow}} \}$. In Fig.~\ref{fig:dino_bb}, we visualize DINO best-buddy pairs extracted between distant frames. As seen, DINO best-buddies provides localized, semantic correspondences across multiple occlusions.

\subsection{Training details}\label{sec:apdx_training_details}
\paragraph{Minibatch sampling.} For memory efficiency, we sample correspondences from a set of 8 frames in each training batch. We sample 512 pairs of optical flow correspondences, at most 1024 pairs of best-buddy features (for $\lossdinobb$ and $\losstunedbb$ separately), and at most 1024 cycle-consistent correspondences. The best-buddy and cycle-consistent correspondences are sampled between 4 pairs of frames. For balanced training, we ensure that 50\% of the optical flow correspondences and 70\% of feature and cycle-consistent correspondences lie in the foreground. We use saliency maps of DINOv2 features~\cite{amir2021deep} for detecting the foreground when ground-truth masks are not available.

\paragraph{Contrastive loss weighting.} As discussed in Sec.~\ref{sec:objective} of the paper, each best-buddy term in $\lossdinobb$ is weighted with a confidence score. For a given pair $\{ \mathdinofeat^{i}, \mathdinofeat^{j} \}$, we measure the confidence score based on 2 metrics: (i) the unimodality of the correlations $\{ \cmap(\bp{p}) = \mathcossim(\mathdinofeat^{i}, \mathfeatt{j}(\bp{p})): \bp{p} \in H' \times W' \}$, (ii) the correlation of the pair $s^{ij} = \cmap(\bp{p}^j)$. To measure (i), we compute the ratio $r_{ij} = s_2 / s_1$, where $s_1 > s_2$ are the 2 highest correlations in $\cmap$. To detect them, we apply non-maximum suppression (NMS) \cite{nms} on the similarity map $\cmap$ with an IoU threshold of 0.2, where we use a box size of 60px for each position. $s_1$ and $s_2$ are, therefore, the top 2 similarities proposed by NMS. Thus, our confidence score is given by $w_{\texttt{dino-bb}}^{ij} = \sigma(a \cdot (1 - \max(r_{ij}, r_{ji})) - b) \cdot 2 (s^{ij})^3 $, where $\sigma(\cdot)$ is the sigmoid function. We fix $a = 27, b = -5.7$ in all our experiments.

For each best-buddy pair $\{ \bp{p}^i, \bp{p}^j \}$ in $\losstunedbb$, we weight the term based on the correlation between the features: $w_{\texttt{rfn-bb}}^{ij} = 2(s^{ij})^3$, where $s^{ij} = \mathcossim(\mathfeatp{i}, \mathfeatp{j})$.

\paragraph{Cycle-consistency loss.} In $\losscyc$ (Eq. \ref{eq:cyc}), for each cycle-consistent pair $\{ \bsx{}^i, \bsx{}^j \}$, we weight the loss term by the cycle-consistency error. Specifically, in Eq. \ref{eq:cyc}, we set $w^{ij}_{\texttt{rfn-cc}} = 0.8^{e_{\texttt{cyc}}}$, where $e_{\texttt{cyc}} = || \bsx{}^i - \mathtracker(\bsx{}^j, i) ||_2$.

\paragraph{Hyperparameters.} We train our model using Adam optimizer \cite{adamopt}, with a learning rate of $0.01$ for all parameters. We decrease the learning rate of the CNN-refiner (Fig. 2) by a factor of $0.999$ every 40 step. For videos of up to 100 frames, the model is trained for 10K iterations. On Kinetics, which contains longer videos (250 frames), we train for 20k iterations. We apply the losses $\losstunedbb$ and $\losscyc$ after 5k training iterations.
The radius $R$ in Eq.~\ref{eq:softargmax} is set to 35px. In $\losstunedbb$ and $\lossdinobb$, we set the temperature $\tau = 0.1$. In $\losscyc$, we use an error threshold $\gamma = 4$. In all our experiments, we use the following weighting in Eq.~\ref{eq:objective}: $\lambda_1 = 25 \times 10^{-5}, \lambda_2 = 5 \times 10^{-5}, \lambda_3 = 0.5, \lambda_4 = 1 \times 10^{-4}$.

\subsection{Architecture}\label{sec:architecture}

\paragraph{Delta-DINO} is a fully convolutional neural network. It comprises 4 layers with channel dimensions of $[3 \rightarrow 64 \rightarrow 128 \rightarrow 256 \rightarrow 1024]$. All layers comprise \texttt{Conv2d} $\rightarrow$ \texttt{BatchNorm2d} $\rightarrow$ \texttt{ReLU} $\rightarrow$ \texttt{BlurPool}, except for the last layer, which comprises \texttt{Conv2d} $\rightarrow$ \texttt{BatchNorm2d}. For \texttt{BlurPool}, we use the antialiased downsampling layers from~\cite{zhang2019shiftinvar}. All convolutional layers have kernel size 5, stride of 1, and reflection padding of 2, except the last layer has reflection padding of 4 and a dilation of 2. To align the residual features with DINO features, we grid-sample from the output of Delta-DINO at the DINO patch-center positions.

\paragraph{CNN-Refiner} \cite{doersch2022tapvid} comprises of \texttt{Conv2d} $\rightarrow$ \texttt{ReLU} $\rightarrow$ \texttt{Conv2d} with channels $[1 \rightarrow 16 \rightarrow 1]$, kernel size 3, and padding 1.

Our model has $\sim$7.6M trainable parameters: $\sim$7.59M for Delta-DINO, $\sim$300 for CNN-Refiner. We use DINOv2-ViTL/14~\cite{oquab2023dinov2} as the DINO backbone in all our experiments. To increase the resolution of DINO features, we modify the stride of the embedding projection layer from 14 to 7~\cite{amir2021deep}.

\subsection{Occlusion Prediction}\label{sec:occ_pred_appendix}

We select the anchor frames $\{ k_i \}$ based on high cos-similarity between query and tracked features: $\{ k_i: \mathcossim(\mathfeatp{}^{k_i}, \mathfeatp{\bp{q}}) \geq 0.7 \}$.
To predict occlusion from trajectory agreement, we calculate an agreement threshold for the trajectory $\mathcal{T}_{\bp{q}}$: for each anchor frame $k$, we sample the median disagreement w.r.t. other anchor frames: $e_k = \med_{k_i}( || \mathtracker(\hat{\bsx{}}^k, k_i) - \hat{\bsx{}}^{k_i} ||_2 )$, and take the maximum of the median errors as the threshold for $\mathcal{T}_{\bp{q}}$: $e_{\bp{q}} = \max_k(e_k)$. A tracked point $\hat{\bsx{}}^t$ is predicted as visible if $\med(d_k) \leq e_{\bp{q}} \ \land \ \mathcossim(\mathfeatp{}^t, \mathfeatp{\bp{q}}) \geq \gamma_{\texttt{occ}} $, where $d_k = || \mathtracker(\bsx{\bp{q}}, k) - \mathtracker(\hat{\bsx{}}^t, k) ||_2$, and $\gamma_{\texttt{occ}} = 0.6$ in all experiments.

\subsection{Ablation Details}\label{sec:ablation_details}

\paragraph{LoRA tuning.}
We use the PEFT implementation \cite{peft} for LoRA. We fine-tune the queries, keys, and values of layers-$\{15, 16\}$ of DINOv2 since we use layer-16 in our tracker (see Tab.~\ref{tab:dino-feature}). We set \texttt{lora\_alpha=0.5, lora\_dropout=0.1, rank=8} when fine-tuning with PEFT.

\paragraph{Raw DINOv2 tracking.}
To track with raw DINOv2 features (see Sec.~\ref{sec:results} of the paper), we use the tracking algorithm described in Sec.~\ref{sec:dino-tracker} and Eq.~\ref{eq:softargmax}, while setting $\mathrefinertensor(\mathframe) = \mathbf{0}$ and $\hmap = \cmap$ (i.e. without Delta-DINO and CNN-Refiner).

\subsection{Benchmarks Evaluation}
\label{sec:apdx_benchmarks}
On the TAP-Vid benchmark we evaluate all methods using "query-strided" sampling, where points on the annotated tracks are sampled as query every five frames \cite{doersch2022tapvid}. All metrics on the TAP-Vid benchmark are computed in 256x256 resolution.
BADJA \cite{biggs2018badja} provides key-point position and visibility labels every 3-5 frames. For evaluation, points are sampled once, at their first visible frame.
For the dotted visualizations shown in Fig.~\ref{fig:comp-tapvid} and the SM, we track a dense grid of points on the query frame, and visualize only tracks that lie on the foreground.

We follow PIPs++ and Co-Tracker's evaluation protocol, and resize frames to their training resolution of 384x512 and 512x896 respectively, before inference. We provide Co-Tracker's query points a support of 6 global and 6 local grid points. TAP-IR and TAP-Net are evaluated at the provided input resolution. For the RAFT baseline, we found that upsamling frames from 256 × 256 improves performance on the TAP-Vid-DAVIS-256 and TAP-Vid-Kinetics-256 benchmarks, and we resize downsampled frames to 480x854 before inference. 
We used Omnimotion's published code to train models for the TAP-Vid-DAVIS-480 and BADJA benchmarks, pre-trained weights were provided for TAP-Vid-DAVIS-256 and TAP-Vid-Kinetics-256.

\section{Complexity}
\label{sec:apdx_complexity}

\subsection{Training time.}
Fitting \dinotracker to a single video with 100 frames takes about 1.6 hours (less than a second per iteration) on a single A100 GPU.
Our training time is $\times\!10$ faster than Omnimotion for the same video.
Training LoRA-tune baseline (see Sec.~\ref{sec:ablations}) with the same settings takes almost 9 hours per video (about 1.5 sec/iteration). This is $\times\!6$ slower than our CNN-based refiner network.

To improve training efficiency, we show that DINO-Tracker can be trained only on a subset of the frames, while still evaluating on \emph{all} frames during inference. Tab. \ref{tab:compressed} reports our performance when training  only $50\%$ or $25\%$ of the frames, thus reducing the training time by the same factor. As seen, our method maintains its performance when trained on $50\%$ and is competitive when trained on $25\%$ of the frames. This demonstrates the ability of our tracker to generalize to \emph{unseen} frames and suggest it can be extended efficiently to longer videos.

\vspace{-0.5cm}
% ------------------------------------------------------------------------ %
% ------------------------ Compression table ------------------------ %
% ------------------------------------------------------------------------ %
\begin{table}[h!]
    \renewcommand{\tabcolsep}{4pt}
    \newcommand{\stdt}[2]{{#1}{\footnotesize$\pm\!{#2}$}}
    \newcommand{\stdtb}[2]{\textbf{#1}{\footnotesize$\pm\!{#2}$}}
    \caption{\emph{Generalization when training on every 2nd and every 4th frame in each video (TAP-Vid-DAVIS-480) on a single A100.}}
% \hspace*{-5mm}
\centering
\resizebox{0.8\linewidth}{!}{
\begin{tabular}{c| c | c | c | c | } 
  
  \hline
  &  $\delta_{avg}^x$ & AJ & OA & train time \\ 
  \hline
  Ours & 80.7 & 65.3 & 88.5 & 90 min. \\
  \hline
  Ours, every 2nd frame & 80.6 & 65.7 & 88.5 & 45 min. \\
  \hline
  Ours, every 4th frame & 79.1 & 64.0 & 87.2 & 22.5 min. \\
 \cline{1-5}
\end{tabular}
 }
\label{tab:compressed}
\end{table}

\newpage
\subsection{Runtime and Memory}

We measure the required compute for \emph{full} inference (both position and visibility) of DINO-Tracker and feed-forward competitors. Tab. \ref{tab:complexity} reports average runtime and allocated memory on TAP-Vid-DAVIS-480 on a single A100
for our tracker and feed-forward methods. Most of our runtime is used for visibility prediction, yet once trained, our total inference time is fastest (note that PIPS++ cannot predict visibility) and is less memory-consuming than TAPIR.

\begin{table}[t!]
    \renewcommand{\tabcolsep}{4pt}
    \newcommand{\stdt}[2]{{#1}{\footnotesize$\pm\!{#2}$}}
    \newcommand{\stdtb}[2]{\textbf{#1}{\footnotesize$\pm\!{#2}$}}
    
\caption{\emph{DAVIS-480 inference time and memory on a single A100.}}
\hspace*{-5mm}
\resizebox{\linewidth}{!}{    
\begin{tabular}{c| a | a | c | c | a | } 
  
  \hline
  & Co-Tracker &  TAPIR & PIPS++ & Ours & Ours  \\ 
  & (full) & (full) & (pos. only) & (pos. only) & (full)  \\ 
  \hline
  Time (sec) & 345.8 &  110.4 &  8.6  &  4.3  & 80.5 \\
  \hline
  GPU-Mem (GB) & 19.2 & 60.0 & 12.0  &  15.2  & 52.6 \\
 \cline{1-6}
\end{tabular}
 }
\label{tab:complexity}
\end{table}

\clearpage

\end{document}